\documentclass{svproc}
\usepackage{url}
\usepackage{xurl}
\usepackage{graphicx}
\usepackage{caption}
\usepackage{subcaption}
\usepackage[colorlinks=true,allcolors=blue]{hyperref}
\usepackage{cite}
\usepackage{multirow}

\begin{document}
\mainmatter              
\title{Gastrointestinal Disease Classification through Explainable and Cost-Sensitive Deep Neural Networks with Supervised Contrastive Learning}
\titlerunning{Gastrointestinal Disease Classification}  
%
\author{Dibya Nath \and  G. M. Shahariar}
\authorrunning{Dibya Nath \and  G. M. Shahariar} 
%
%
\institute{Ahsanullah University of Science and Technology, Dhaka, Bangladesh\\
 \email{dibya.nath8420@gmail.com, sshibli745@gmail.com}
}

\maketitle              

\begin{abstract}
Gastrointestinal diseases pose significant healthcare chall-enges as they manifest in diverse ways and can lead to potential complications. Ensuring precise and timely classification of these diseases is pivotal in guiding treatment choices and enhancing patient outcomes. This paper introduces a novel approach on classifying gastrointestinal diseases by leveraging cost-sensitive pre-trained deep convolutional neural network (CNN) architectures with supervised contrastive learning. Our approach enables the network to learn representations that capture vital disease-related features, while also considering the relationships of similarity between samples. To tackle the challenges posed by imbalanced datasets and the cost-sensitive nature of misclassification errors in healthcare, we incorporate cost-sensitive learning. By assigning distinct costs to misclassifications based on the disease class, we prioritize accurate classification of critical conditions. Furthermore, we enhance the interpretability of our model by integrating gradient-based techniques from explainable artificial intelligence (AI). This inclusion provides valuable insights into the decision-making process of the network, aiding in understanding the features that contribute to disease classification. To assess the effectiveness of our proposed approach, we perform extensive experiments on a comprehensive gastrointestinal disease dataset, such as the Hyper-Kvasir dataset. Through thorough comparisons with existing works, we demonstrate the strong classification accuracy, robustness and interpretability of our model. We have made the implementation of our proposed approach publicly available\footnote{ \url{https://github.com/dibya404/Gastrointestinal-Disease-Classification-through-Explainable-and-Cost-Sensitive-DNN-with-SCL}}.

\end{abstract}

\section{Introduction}
The gastrointestinal (GI) tract is prone to numerous diseases and anomalies that can significantly impact an individual's health and well being. The International Agency for Research on Cancer has reported that the incidence of gastrointestinal cancer globally in 2018 was estimated at 4.8 million new cases, which accounted for 26\% of all cancer cases worldwide and the number of deaths associated with gastrointestinal cancer was estimated to be 35\% of all cancer-related deaths globally \cite{r3}.  Certain conditions can be life-threatening and require prompt detection and intervention to increase the chances of successful treatment and full recovery. But it is unfortunate that certain gastrointestinal diseases pose challenges in their detection or may create ambiguity during medical screening due to the presence of image noise that masks important details \cite{r4}.
\par Endoscopy and colonoscopy are currently the standard procedures for performing investigations of the GI tract. Endoscopy and colonoscopy are intrusive procedures that involve the insertion of a tube, with a small camera attached to it, either orally or rectally which requires thorough examination by a doctor to inspect the interior of the GI tract organs and investigate for any anomalies such as polyps (small, benign growths), ulcers or cancer. The ability to identify polyps influences the likelihood of developing cancer. Up to 20\% of the polyps can be undetected during examination \cite{r1}. Furthermore, the subjective interpretation of endoscopy images can be time-intensive and minimally repetitive, potentially leading to an inaccurate diagnosis \cite{r4}. Given the potential for inaccuracies in diagnosis by human professionals, as a result of human error and other limitations, the utilization of a computer-aided automated system could be advantageous in detecting gastrointestinal polyps in their early stages of cancer. Deep learning techniques, the representation of an alternative form of machine learning methods, have been applied in multiple aspects of gastrointestinal endoscopy, that include colorectal polyp detection and classification, analysis of endoscopic images for diagnosis of helicobacter pylori infection detection and depth assessment of early gastric cancer, and detection of various abnormalities in wireless capsule endoscopy images \cite{r8, r9, r10}.
\par In this paper, we propose cost-sensitive pre-trained convolutional neural network (CNN) architectures (EfficientNet, DenseNet, ResNet, Xception, Inception-v3) which are fine-tuned using supervised contrastive training objective for diagnosing gastrointestinal diseases on 23 distinct classes. No prior studies have explored this combination in the context of gastrointestinal disease detection. Supervised contrastive learning \cite{khosla2020supervised} is a training approach that has shown improved performance compared to traditional supervised training using cross-entropy loss in classification tasks. This method involves a two-stage process for training an image classification model. Firstly, an encoder is trained to generate vector representations of input images in a way that images from the same class have higher similarity in their vector representations compared to images from different classes. Secondly, the learned representations are utilized for downstream tasks by fine-tuning a classifier while keeping the encoder frozen. To enhance the training process of our proposed approach, we employed cost-sensitive learning, which considers the costs associated with prediction errors and other potential costs when training the classifier model. The lack of transparency and interpretability in modern AI systems can be especially concerning in domains like healthcare, where AI-driven decisions can significantly impact human lives and well-being. Thus, we leveraged explainable artificial intelligence (XAI) techniques such as GradCAM, GradCAM++, faster version of ScoreCAM, LayerCAM to provide visual explanations for the classifier's decisions regarding classifying input images. 
\par The rest of the paper is organized as follows: section \ref{rel} presents some of the related works, section \ref{bs} contains a brief overview of the pre-trained models and working procedures of the cost-sensitive, contrastive learning and explainable AI techniques, the proposed methodology is detailed in section \ref{pm}, performance evaluation of the proposed approach is reported in section \ref{eval}, section \ref{xai} illustrates the interpretation of the proposed approach and finally, section \ref{conclu} concludes the study with some future research directions.

\section{Related Works}\label{rel}
Several prior works have explored the utilization of pre-trained convolutional neural network (CNN) architectures and transfer learning. Below, we provide a discussion of some of these relevant studies.
For polyp detection, localization, and segmentation Jha et al. \cite{r13} evaluated several state-of-the-art methods on Kvasir-SEG dataset. For the detection and localization, the author proposed ColonSegNet, which displayed a comparison between  mean IoU (0.81) and average precision (0.80), accompanied with 180 fps speed. ColonSegNet reached a dice coefficient of 0.8206. For the segmentation, the best average speed was 182.38 fps. 
Khan et al.\cite{r14} utilized a Private Stomach dataset comprising ulcer, bleeding, and healthy classes. They fine-tuned Inception-v3, DenseNet201 and AlexNet, and transformed them into SecureCNN models for feature extraction from the images and feature optimization of the network. The outcome of feature networks optimization without fine-tuning manifested as a classification accuracy,  a false negative rate of respectively 92.4\% and  7.6\%. Upon using optimized features of fine-tuned networks, the classification accuracy improved by 4.76\% to 96.8\%, where the false negative rate was 3.2\%.
Sun et al\cite{r15} used their own dataset of gastroscopic images and applied the DenseNet-121 network as the backbone on the ImageNet dataset and applied the  parameters that were pre-trained before the fully-connected layers. They added a non-local block into DenseNet and through extensive experimentation they achieved an accuracy, a recall and an F1-score of respectively 96.79\%, 94.92\%, and 94.70\% utilizing proposed improved DenseNet. The authors compared their method with Xception, Inception, ResNet-101, and DenseNet-121 models to demonstrate the superior achievement of the introduced method. 
Shin et al. \cite{r16} employed polyp-frame datasets that are publicly available, and two colonoscopy video databases to apply a region-based CNN  model, using Inception ResNet model as transfer learning with the purpose of identifying polyps automatically in colonoscopy images and videos. As the dataset is small and the detection task is challenging, four different image augmentation strategies were designed and examined for training deep networks.  The Aug-I based method correctly detects 3137 out of 3856 polyps, but the false positives number was 1145, resulting in a recall, a precision of respectively 81.4\% and 73.3\%.
For polyp detection purposes Ribeiro et al.\cite{r17} performed binary classification  and applied a transfer learning approach wherein the selected architectures were: ResNet-34, ResNet-50, ResNet-152, MobileNetV2, and VGG16\_bn. Upon observing the performance, ResNet-152 was elected superior with 96.60\% accuracy, 94.30\% precision,  94.30\% recall and 0.90 f1 score.

\section{Background Study}\label{bs}
\textbf{(a) Pre-trained Convolutional Neural Networks:} Convolutional Neural Networks (CNNs) are a type of artificial neural network outlined to handle data such as images or videos, that incorporates a grid-like structure. They consist of several layers that perform a series of operations on the input data, including convolution, pooling, and activation functions. These layers work together to extricate and learn pertinent features from the input data, allowing the network to recognize complex objects and patterns.\\
\noindent\textbf{ResNet50}: Residual Network (ResNet) is a deep learning architecture proposed by He et al. \cite{r23}. One of the major advancements introduced by ResNet is the incorporation of skip connections. These skip connections enable the network to bypass certain layers and directly connect earlier layers to later ones, allowing for the flow of information to traverse more easily. ResNet50, a specific variant of ResNet, is particularly notable as it comprises 50 layers within the architecture. This specific configuration of ResNet50 has gained significant popularity and has been extensively utilized for various image classification and computer vision tasks. \\
\noindent\textbf{EfficientNet}: EfficientNet is a deep learning architecture designed for image classification tasks, which was introduced by Tan and Le \cite{r24}. The authors proposed a novel approach called compound scaling, which aims to balance the depth, width, and resolution factors of convolutional neural networks (CNNs). By carefully scaling the different dimensions of the network, EfficientNet achieves an optimal balance between model size and computational efficiency, resulting in highly effective image classification capabilities.\\
\noindent\textbf{DenseNet}: Dense Convolutional Network (DenseNet) was proposed by Huang et al. \cite{r25}. The assessment of their proposed architecture on CIFAR-10, CIFAR-100, SVHN, and ImageNet revealed DenseNets as out-performer among the existing state-of-the-art models on most of these benchmarks while achieving high accuracy with reduced computational requirements. The primary innovation of DenseNet is its use of dense $L(L+1)/2$ direct connectivity between layers where in traditional CNN, $L$ layers have $L$ connections.\\
\noindent\textbf{Inception}: Inception was introduced by Szegedy et al. \cite{r26}.The key innovation of Inception is the use of a multi-branch architecture allowing the network to capture features using filters of different sizes at different scales. Other deep learning models, such as Inception-v3 were derived from Inception architecture.\\
\noindent\textbf{Xception}: 
Xception is a deep learning architecture that was introduced by François Chollet \cite{r27}. The key innovation of Xception lies in its utilization of depthwise separable convolutions, which differ from traditional convolutions. By employing this novel convolutional technique, Xception achieves improved efficiency and effectiveness in capturing spatial and channel-wise dependencies within the network, leading to enhanced classification capabilities.\\
\noindent \textbf{(b) Supervised Contrastive Learning:} 
The primary objective of supervised contrastive learning \cite{khosla2020supervised} is to bring images belonging to the same class closer together in the embedding space, while simultaneously creating greater separation between images of different classes. By doing so, it enables the model to learn meaningful and compact representations that capture high-level semantic information. Contrastive learning relies on the use of positive and negative pairs of images. Positive pairs are formed by taking two augmented variations of an identical image, whereas negative pairs are composed of images belonging to distinct classes. The aim is to maximize the separation between negative pairs and minimize the proximity between positive pairs. To encourage the model to learn robust and invariant representations, strong data augmentation strategies such as random cropping, rotation, color jittering are applied to both images in a pair to capture diverse perspectives of the same object. A similarity metric, such as cosine similarity or Euclidean distance, is employed to quantify the similarity between image representations in the embedding space. One of the main components of supervised contrastive learning is the contrastive training objective. The aim is to enhance similarity between positive pairs of images and decrease similarity between negative pairs in order to improve the quality of image embeddings. several training objectives, such as the TripletMargin loss, MaxMargin loss, NPairs loss, and infoNCE loss can be used to achieve this objective.\\
\noindent \textbf{(c) Cost-Sensitive Learning:} Cost-sensitive learning refers to a subfield of machine learning that involves incorporating the costs associated with prediction errors while training a machine learning model. This is an area closely linked to imbalanced learning, which deals with classification in datasets where the class distribution is heterogeneous. The primary objective of cost-sensitive learning for imbalanced classification is to assign various costs to the sort of misclassification errors that might occur. Machine learning algorithms designed for classification largely assume that the number of examples contained in each observed class is balanced. However, in real-world scenarios, a dataset that is known as an imbalanced classification problem, may exhibit skewed class distributions. The assumption of uniformity of the costs related with misclassification, such as false positives and false negatives are also often made by the majority of classifiers. A notable example is the diagnosis of cancer, where misclassification of a cancer case as negative carries far more severe consequences than a false positive diagnosis. Because indeed, patients may suffer and lose their lives due to delays in treatment resulting from a missed diagnosis.\\
\noindent \textbf{(d) Explainable Artificial Intelligence (XAI):} Conventional machine learning models and algorithms often function in a manner that makes it challenging to comprehend the underlying process by which they generate predictions, decisions, or recommendations. In the past few years, there has been a notable increase in the prevalence of intricate decision systems, such as Deep Neural Networks (DNNs), which exhibit a high level of opacity and complexity. The reason behind the success of these models is their large parameter space and efficient learning algorithms, which comprise multiple layers and millions of parameters. Such complexity has made them known as complex black box models that are difficult to interpret. As the use of these black box machine learning models becomes more common in critical scenarios, stakeholders are demanding transparency in their decision-making process. To balance between the high learning performance of current AI systems and the need for transparency, the field of explainable AI (XAI) has emerged. In the field of image processing, two commonly employed explainable artificial intelligence (XAI) techniques are gradient-based methods like GradCAM \cite{gradcam}, GradCAM++ \cite{gradcamplusplus}, faster version of ScoreCAM \cite{scorecam}, LayerCAM \cite{layercam} and perturbation-based methods like LIME \cite{lime}. These techniques offer valuable insights into the inner workings of image processing models, aiding in the interpretability and understanding of their predictions.

\begin{figure}[h]
         \centering
         \includegraphics[width= 125mm]{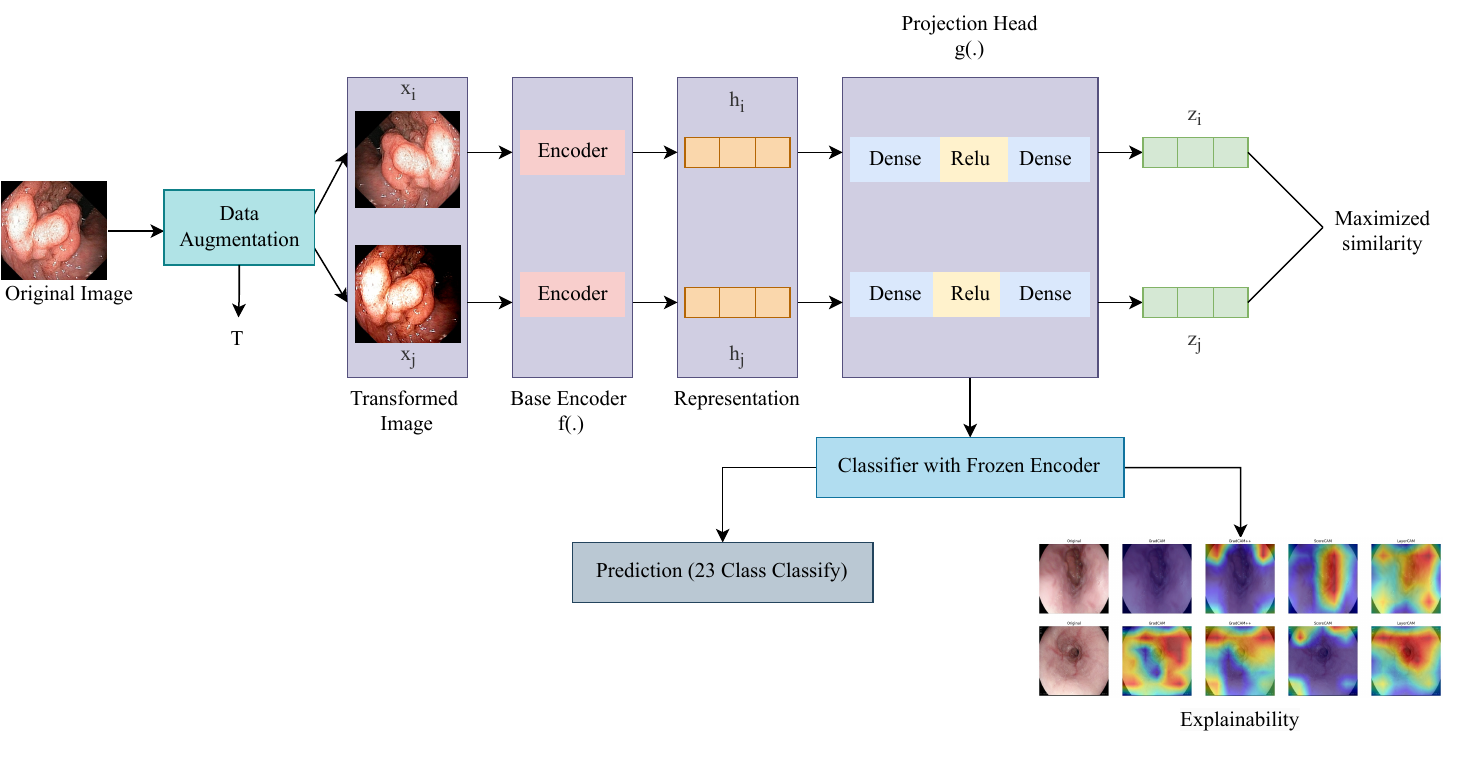}
         \caption{Schematic diagram of the proposed methodology.}
         \label{fig:Proposed Methodology}
\end{figure}

\section{Proposed Methodology}\label{pm}
Figure \ref{fig:Proposed Methodology}showcases the schematic representation of our proposed methodology. We describe each step below in details.\\
\noindent \textbf{Step 1) Data augmentation}: Initially, we employed various data augmentation techniques to augment the training data and introduce more diversity. These techniques included random cropping, flipping, color dropping, and color jittering. By applying these augmentation strategies, we aimed to enhance the variability in the training dataset, allowing the model to learn robust and generalized representations of the data.\\
\noindent \textbf{Step 2) Image processing}: During this step, we standardized the dimensions of all the images by resizing them to a resolution of 224 × 224. This resizing process ensured that all the images had consistent dimensions, enabling compatibility and uniformity throughout the dataset. By reshaping the images to this specific resolution, we prepared them for further processing and analysis in subsequent steps of our workflow.\\
\noindent\textbf{Step 3) Contrastive training}: We trained a pre-trained CNN architecture (Xception which performed best in our study) using MaxMargin contrastive loss. The pre-trained CNN architecture served as the base encoder in our approach. On top of the encoder, we added two dense layers along with the ReLU activation function to form the projection head. The objective was to minimize the distance between images belonging to the same class within a batch, while maximizing otherwise.\\
\noindent \textbf{Step 4) Classifier with frozen encoder}: The encoder model was specifically designed to take an image as input and produce a feature vector with multiple dimensions. On top of the frozen encoder, we constructed a classifier comprising a fully-connected layer and a softmax layer for the target classes. During fine-tuning, the classifier was trained using cross-entropy loss while keeping the weights of the trained encoder frozen. Only the weights of the fully-connected layers with softmax were optimized in this process. We fine-tune the classifier model two ways: (a) without cost-sensitive learning and (b) with cost-sensitive learning. To automatically compute the class weights in cost-sensitive learning we used sklearn\footnote{\url{https://scikit-learn.org/stable/modules/generated/sklearn.utils.class_weight.compute_class_weight.html}} library.\\
\noindent \textbf{Step 5) Interpretation using XAI}: In this particular stage, we employed four XAI techniques, namely GradCAM, GradCAM++, faster version of ScoreCAM, and LayerCAM, to visually illustrate the predictions made by the classifier. These techniques were utilized to provide insights into why the classifier attributed a specific input test image to a particular class. By utilizing these XAI methods, we aimed to gain a better understanding of the decision-making process and provide visual explanations for the classifier's assigned classifications.

We repeated all the steps for twelve different pre-trained CNN models including EfficientNetB0, EfficientNetB1, EfficientNetB2, EfficientNetV2B0, EfficientNetV2B1, EfficientNetV2B2, Inception-v3, Xception, ResNet50, DenseNet121, DenseNet169.\\

\section{Evaluation}\label{eval}
In this section, we evaluate our proposed approach in terms of performance analysis. 
\subsection{Dataset Description}
The HyperKvasir dataset \cite{r11} is a comprehensive collection of high-quality gastrointestinal (GI) endoscopy images and videos, designed to aid medical image analysis research. It comprises a diverse range of images, including normal and abnormal GI tract conditions, captured using different types of endoscopes. The dataset contains annotations such as lesion location and categorical labels for various abnormalities, and with thousands of images and videos, it serves as a valuable resource for tasks like lesion detection, classification, segmentation, and disease localization. The dataset promotes collaboration and supports the development of AI-based solutions for automated analysis in GI endoscopy, which can assist in the diagnosis and treatment of GI disorders. The dataset consists of a total of 110,079 images in JPEG format, out of which 10,662 images are labeled into 23 distinct classes. The images are in RGB format.

\begin{figure}[h]
         \centering
         \includegraphics[scale=0.27]{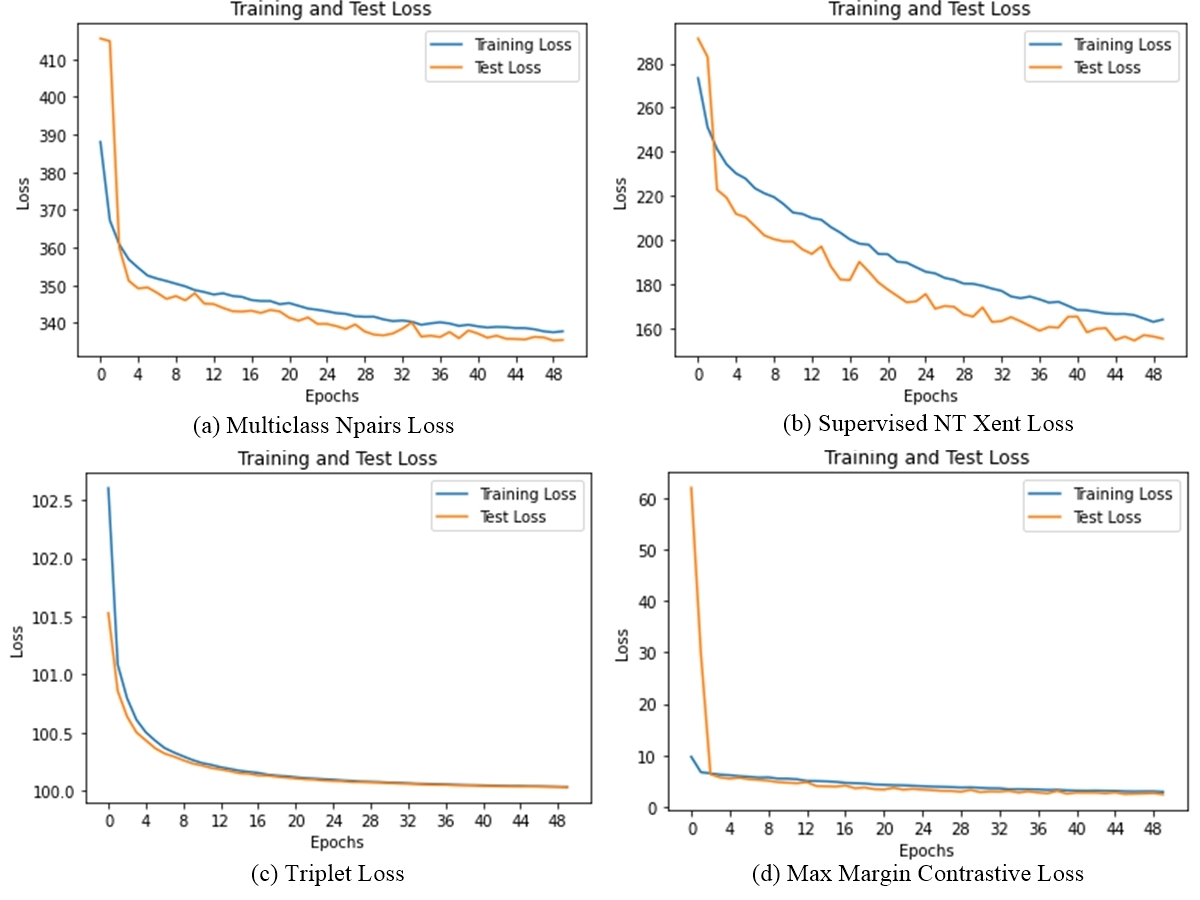}
         \caption{Four different contrastive training and test loss curves for ResNet50 model.}
         \label{fig:Loss Curve}
\end{figure}

\subsection{Experiments}
In this study, we carried out the following two experiments:\\
\noindent\textbf{(a) Contrastive learning}: We investigated whether training pre-trained CNN architectures using supervised contrastive learning before fine-tuning them for the classification task would result in improved performance.\\
\noindent\textbf{(b) Contrastive then cost-sensitive learning}: We explored whether the pre-trained CNN architecture, trained using contrastive learning, could be further fine-tuned using cost-sensitive cross-entropy training to achieve enhanced classification performance.

\begin{table}[h!]
\centering
\caption{Performance comparison of different models using contrastive and cos-sensitive learning among themselves and with some existing works.}
\label{tab:Accuracy of Differnt Models}
\resizebox{\textwidth}{!}
{\begin{tabular}{ccccc} 
\hline
\begin{tabular}[c]{@{}c@{}}\textbf{Training }\\\textbf{~Objective~}\end{tabular}    & \textbf{Model~}                                                                       & \begin{tabular}[c]{@{}c@{}}\textbf{Test }\\\textbf{~ ~Accuracy~ ~}\end{tabular} & \begin{tabular}[c]{@{}c@{}}~ ~\textbf{MaxMargin~~}\\\textbf{Loss}\end{tabular} & \begin{tabular}[c]{@{}c@{}}~ \textbf{Weighted avg~~}\\\textbf{F1-score}\end{tabular}  \\ 
\hline
\multirow{11}{*}{\begin{tabular}[c]{@{}c@{}}Contrastive \\Learning\end{tabular}}    & EfficientNetB0                                                                        & 63.61                                                                           & 0.045                                                                          & 0.56                                                                                  \\
                                                                                    & EfficientNetB1                                                                        & 54.899                                                                          & 0.055                                                                          & 0.47                                                                                  \\
                                                                                    & EfficientNetB2                                                                        & 60.196                                                                          & 0.056                                                                          & 0.50                                                                                  \\
                                                                                    & ~ ~ ~EfficientNetV2B0~ ~ ~                                                            & 71.96                                                                           & 0.0335                                                                         & 0.66                                                                                  \\
                                                                                    & ~ EfficientNetV2B1~~                                                                  & 69.432                                                                          & 0.03                                                                           & 0.62                                                                                  \\
                                                                                    & ~ EfficientNetV2B2~~                                                                  & 68.21                                                                           & 0.047                                                                          & 0.61                                                                                  \\
                                                                                    & ResNet50                                                                              & 79.18                                                                           & 0.024                                                                          & 0.76                                                                                  \\
                                                                                    & DenseNet121                                                                           & 79.6                                                                            & 0.335                                                                          & 0.75                                                                                  \\
                                                                                    & DenseNet169                                                                           & 82.46                                                                           & 0.026                                                                          & 0.83                                                                                  \\
                                                                                    & InceptionV3                                                                           & 76.27                                                                           & 0.027                                                                          & 0.07                                                                                  \\
                                                                                    & Xception                                                                              & 83.544                                                                          & 0.027                                                                          & 0.81                                                                                  \\ 
\hline
\begin{tabular}[c]{@{}c@{}}Contrastive \\Learning\\ +\\ Cost-Sensitive\end{tabular} & Xception                                                                              & \textbf{88.74}                                                                  & 0.027                                                                          & \textbf{0.86}                                                                         \\ 
\hline
                                                                                    & \begin{tabular}[c]{@{}c@{}}ResNet50~\\EfficientNetB7 \cite{9850571}\end{tabular}                 & 88.18                                                                           &                                                                                &                                                                                       \\ 
\hline
                                                                                    & \begin{tabular}[c]{@{}c@{}}EfficientNet \cite{9474707}\\(teacher student framework)\end{tabular} & 84                                                                              &                                                                                & 0.84                                                                                  \\
\hline
\end{tabular}}
\end{table}

\subsection{Hyper-parameter settings} All experiments were conducted using a 12 GB NVIDIA Tesla K80 GPU on Google Colaboratory. The implementations were carried out using the TensorFlow framework. A batch size of 64, AdamW optimizer, learning rate of 2e-5, and softmax activation function were used for all the experiments. The encoder in the contrastive learning phase was trained for 50 epochs. Due to limited availability of images in certain classes of the original dataset, data augmentation techniques were employed to address the class imbalance during contrastive training. The classifier, with the encoder frozen, was trained for 220 epochs using categorical cross-entropy loss. For experimental purposes, the data was divided into an 80\% training set and a 20\% testing set. All images were resized to a dimension of $224 \times 224$.

\subsection{Evaluation metrics} We utilized a variety of evaluation metrics to assess the performance of the models in our study. Accuracy was employed to determine the overall correctness of the model's predictions by comparing the number of correct predictions to the total number of predictions made. Precision was used to evaluate the model's ability to correctly identify positive instances among all instances predicted as positive, thereby aiding in the identification of false positives. Recall, also known as sensitivity, measured the model's capability to accurately detect positive instances among all actual positive instances. The F1 Score, a balanced measure derived from the harmonic mean of precision and recall, provided a comprehensive evaluation of the model's overall performance, considering both metrics simultaneously. These evaluation metrics collectively contributed to a thorough assessment of the models' performance in our study.

\begin{figure}
         \centering
         \includegraphics[scale=0.6]{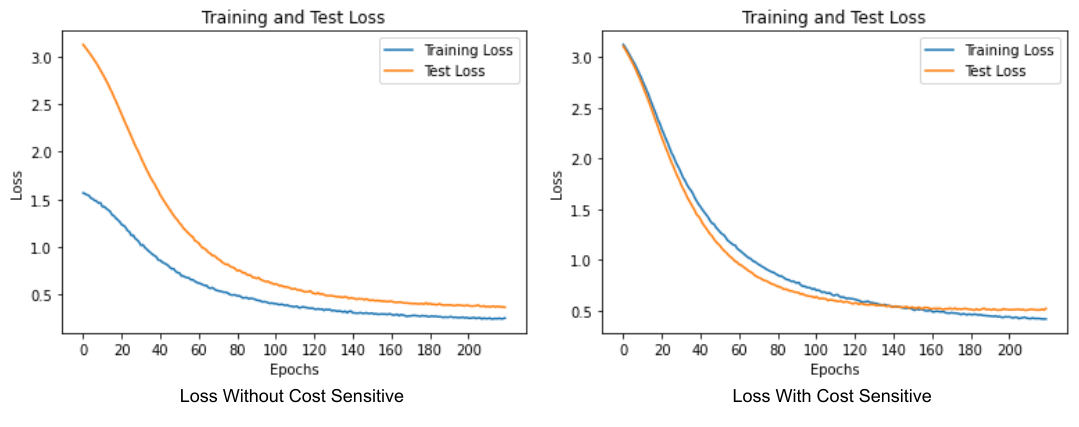}
         \caption{Training and test loss graph for Xception model with and without cost-sensitive learning.}
         \label{fig:Cost sensitive Accuracy}
\end{figure}

\subsection{Experimental results \& analysis}
This section presents the outcomes of the conducted experiments.\\
\textbf{(a) Contrastive learning:} Initially, we employed four contrastive loss functions: MaxMargin loss, TripletMargin loss, NPairs loss, and supervised NT-XENT loss to train ResNet50 as the encoder during contrastive training. Our analysis revealed that ResNet50 exhibited superior performance in terms of train and test loss when trained with MaxMargin loss. Figure \ref{fig:Loss Curve} illustrates the training and test loss curves for ResNet50 with the four different contrastive loss functions. Based on these results, we selected MaxMargin loss as the preferred loss function for training all the twelve pre-trained CNN architectures during contrastive training. It is evident from Table \ref{tab:Accuracy of Differnt Models} that Xception model achieved the best test accuracy of around 84\%. EfficientNetB1, EfficientNetB2 and EfficientNetB0 yielded the least satisfactory results with test accuracies of 54.899\%, 60.196\% and 63.61\% respectively. While EfficientNetV2B2, EfficientNetV2B1 and EfficientNetV2B0 exhibited slightly better outcomes, their test accuracies were not in close proximity to the highest level of accuracy attained, with values of 68.21\%, 69.432\% and 71.96\% respectively. Inception-v3 performed better than the previously mentioned models with a test accuracy of 76.27\% but still fell short of the optimal performance level. The test accuracy of DenseNet121 was 79.6\%, which was superior to most other models. However, the encoder test loss was significantly higher, recording 0.335\%. In contrast, ResNet50 exhibited a test accuracy of 79.18\% and an encoder test loss of 0.024\%.

\begin{table}[h]
\centering
\caption{Classification report for Xception model with and without cost-sensitive learning.}
\label{tab:Classification Report Without Cost-Effective}
\resizebox{\textwidth}{!}
{\begin{tabular}{ccccccc} 
\cline{2-7}
\multicolumn{1}{l}{}             & \multicolumn{3}{c}{\textbf{Without cost-sensitive}}        & \multicolumn{3}{c}{\textbf{With cost-sensitive}}            \\ 
\hline
\textbf{Class Name}              & \textbf{Precision} & \textbf{Recall} & \textbf{F1 - Score} & \textbf{Precision} & \textbf{Recall} & \textbf{F1 - Score}  \\ 
\hline
cecum                            & 0.94               & 0.99            & 0.96                & 0.95               & 1.00            & 0.97                 \\
ileum                            & 0.00               & 0.00            & 0.00                & 0.00               & 0.00            & 0.00                 \\
retroflex\_rectum                & 0.91               & 0.91            & 0.91                & 0.91               & 0.99            & 0.94                 \\
hemorrhoids                      & 0.00               & 0.00            & 0.00                & 0.00               & 0.00            & 0.00                 \\
polyps                           & 0.96               & 0.99            & 0.98                & 0.99               & 1.00            & 0.99                 \\
ulcerative\_colitis\_grade\_0\_1 & 0.00               & 0.00            & 0.00                & 0.00               & 0.00            & 0.00                 \\
ulcerative\_colitis\_grade\_1    & 0.00               & 0.00            & 0.00                & 0.00               & 0.00            & 0.00                 \\
ulcerative\_colitis\_grade\_1\_2 & 0.00               & 0.00            & 0.00                & 0.00               & 0.00            & 0.00                 \\
ulcerative\_colitis\_grade\_2    & 0.52               & 0.76            & 0.62                & 0.56               & 0.94            & 0.70                 \\
ulcerative\_colitis\_grade\_2\_3 & 0.00               & 0.00            & 0.00                & 0.00               & 0.00            & 0.00                 \\
ulcerative\_colitis\_grade\_3    & 0.00               & 0.00            & 0.00                & 0.00               & 0.00            & 0.00                 \\
bbps\_0\_1                       & 0.94               & 0.96            & 0.95                & 0.96               & 0.98            & 0.97                 \\
bbps\_2\_3                       & 0.98               & 0.99            & 0.98                & 1.00               & 1.00            & 1.00                 \\
impacted\_stool                  & 0.94               & 0.71            & 0.81                & 1.00               & 0.85            & 0.92                 \\
dyed\_lifted\_polyps             & 0.80               & 0.83            & 0.81                & 0.95               & 0.93            & 0.94                 \\
dyed\_resection\_margins         & 0.83               & 0.81            & 0.82                & 0.94               & 0.96            & 0.95                 \\
pylorus                          & 0.87               & 0.88            & 0.92                & 0.93               & 1.00            & 0.97                 \\
retroflex\_stomach               & 0.95               & 0.99            & 0.97                & 0.99               & 0.99            & 0.99                 \\
z\_line                          & 0.68               & 0.85            & 0.75                & 0.71               & 0.94            & 0.81                 \\
barretts                         & 0.00               & 0.00            & 0.00                & 0.00               & 0.00            & 0.00                 \\
barretts\_short\_segment         & 0.00               & 0.00            & 0.00                & 0.00               & 0.00            & 0.00                 \\
esophagitis\_a                   & 0.33               & 0.22            & 0.26                & 0.55               & 0.21            & 0.30                 \\
esophagitis\_b\_d                & 0.45               & 0.23            & 0.31                & 0.54               & 0.52            & 0.53                 \\ 
\hline
\textbf{Macro avg}               & 0.48               & 0.49            & 0.48                & 0.52               & 0.53            & 0.52                 \\
\textbf{Weighted avg}            & 0.80               & 0.84            & 0.81                & 0.85               & 0.89            & 0.86                 \\
\hline
\end{tabular}}
\end{table}

\noindent \textbf{(b) Contrastive then cost-sensitive learning:} Based on the previous experiment, where we observed that the Xception model trained with contrastive learning achieved the highest test accuracy, we proceeded to investigate whether incorporating cost-sensitive learning during the fine-tuning phase of Xception as a classifier could further enhance its classification performance. The results presented in Table \ref{tab:Accuracy of Differnt Models} demonstrate that the Xception model consistently outperformed the other models, achieving a classifier test accuracy of approximately 89\%. This finding suggests that, given the imbalanced nature of the dataset with 23 classes, assigning appropriate costs to incorrect predictions can greatly benefit the classifier's performance.

In Figure \ref{fig:Cost sensitive Accuracy}, we present the graph depicting the train and test loss of Xception with and without the utilization of cost-sensitive learning. Table \ref{tab:Classification Report Without Cost-Effective} displays the precision, recall, and F1-scores for each class of Xception, both with and without the incorporation of cost-sensitive learning. These visualizations and metrics provide valuable insights into the performance and impact of cost-sensitive learning on the Xception model. When cost-sensitive learning was not employed for the Xception model, the weighted F1-score achieved was 81\%, with precision of 80\% and recall of 84\%. It yielded macro-average precision, recall and F1-score of 0.48, 0.49 and 0.48 respectively. These results indicate that while the model is successful in identifying a large portion of positive instances, it may also generate more false positives. Despite using equal class weights, the outcomes were not satisfactory. Figure \ref{fig:Cost sensitive Accuracy} visually depicts the accuracy of the Xception model throughout training, illustrating minimal fluctuations until convergence was reached after 60 epochs. Conversely, the Xception model with the cost-sensitive learning produced macro-average precision, recall and F1-score of 0.52, 0.53, and 0.52, respectively and weighted average precision, recall and F1-score of 0.85, 0.89 and 0.86 respectively. We can deduce that the use of the cost-sensitive learning in fine-tuning the Xception model leads to a superior classification performance. Our approach using Xception model outperformed two existing works as reported in Table \ref{tab:Accuracy of Differnt Models}.

We believe that the integration of supervised contrastive learning and cost-sensitive learning enhances feature representation, tackles challenges related to class imbalance and misclassification costs, takes into account context-specific costs, and improves the interpretability and transparency of the methodology. As a result, we observe better outcomes in the detection of gastrointestinal diseases.

\section{Model interpretation using XAI}\label{xai}
\begin{figure}[htbp]
         \centering
         \includegraphics[scale=0.25]{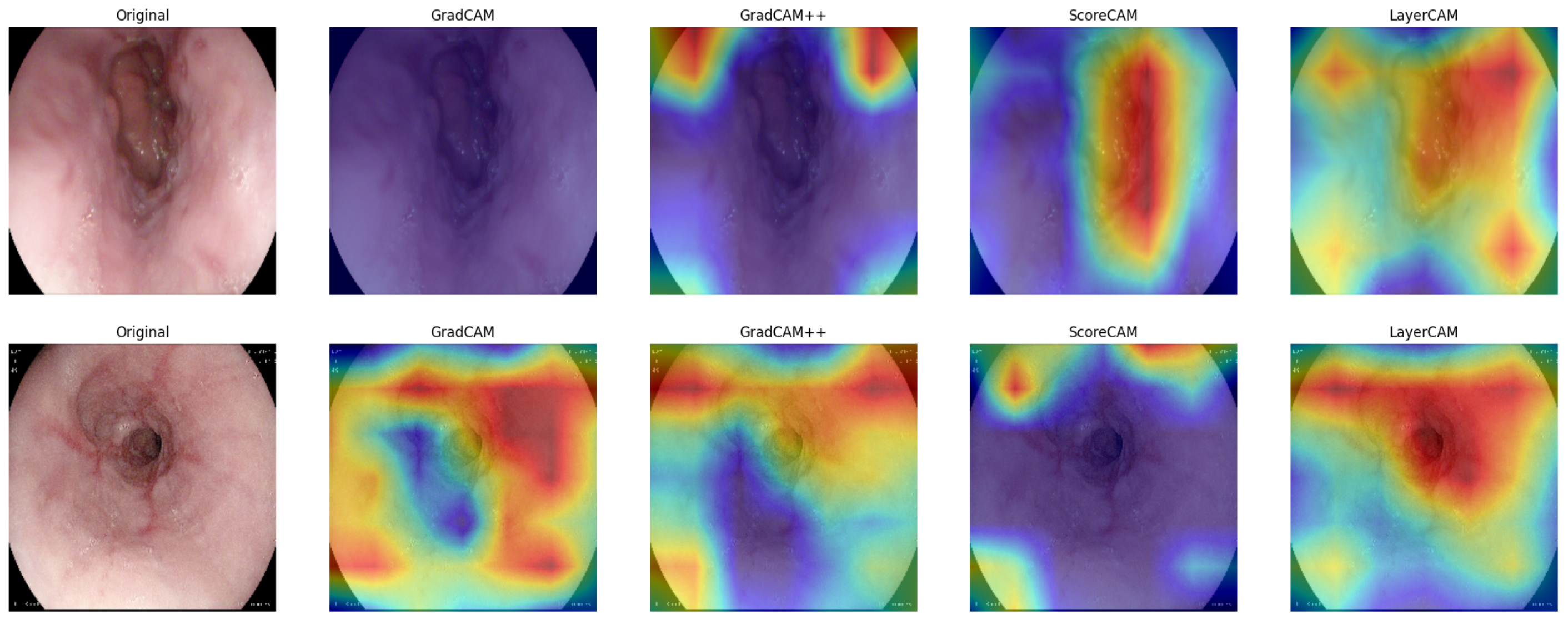}
         \caption{Two sample outputs of correctly classified instances by Xception with cost-sensitive learning using XAI techniques.}
         \label{fig:r2}
\end{figure}

\begin{figure}[htbp]
         \centering
         \includegraphics[scale=0.25]{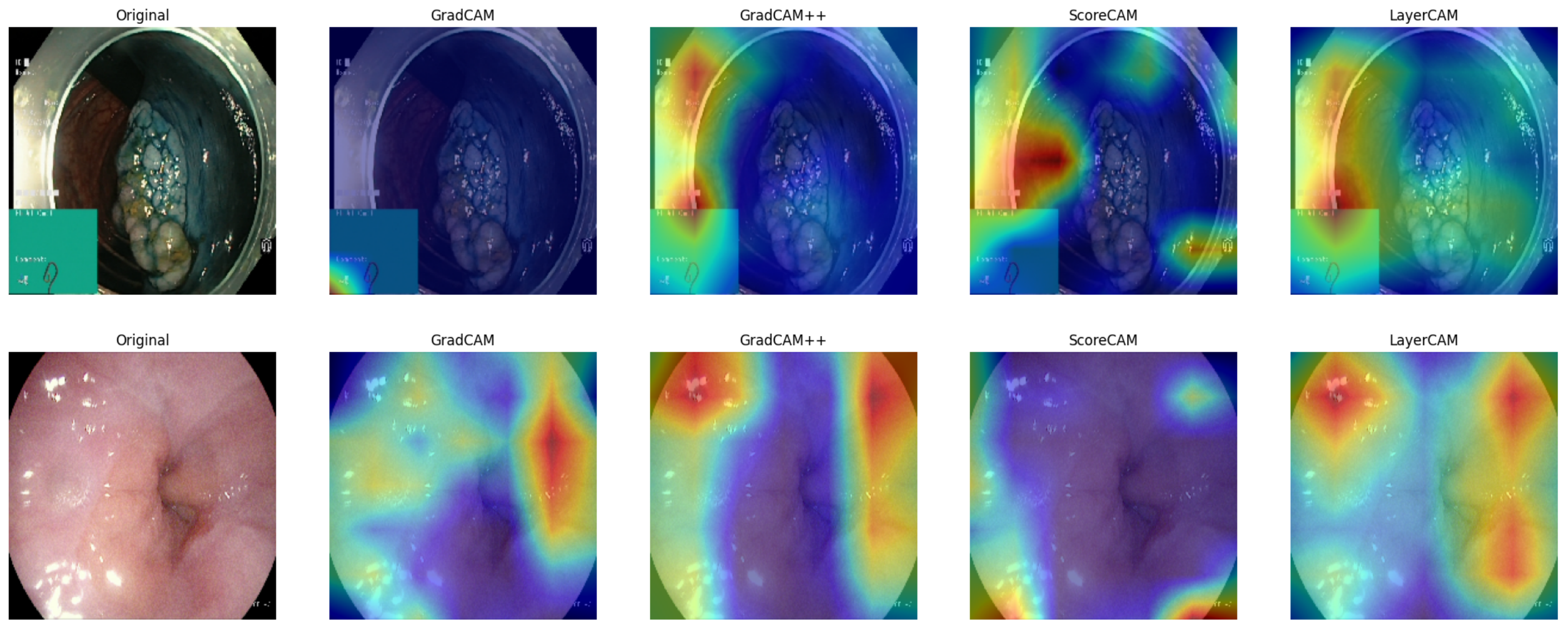}
         \caption{Two sample outputs of misclassified instances by Xception with cost-sensitive learning using XAI techniques.}
         \label{fig:r4}
\end{figure}

We utilized a neural network visualization toolkit available at \footnote{\url{https://github.com/keisen/tf-keras-vis}} to implement GradCAM, GradCAM++, a faster version of ScoreCAM, and LayerCAM for visualizing the correct and incorrect classifications made by the Xception model with and without cost-sensitive learning. In Figure \ref{fig:r2}, we present two images that were correctly classified, accompanied by the corresponding XAI explanations using the four visualization techniques. Figure \ref{fig:r4} showcases two misclassified images, again with the corresponding XAI explanations. The original image is displayed on the left side, while the XAI visualizations are presented on the right.

Figure \ref{fig:r2} provides visualizations of the image explanations using different XAI techniques. Both GradCAM and GradCAM++ show certain areas of interest, but they may not provide sufficient information to fully explain the model's behavior. On the other hand, Faster ScoreCAM effectively highlights the focal points of the image, indicating the model's ability to identify relevant regions of interest. Notably, LayerCAM offers a more realistic explanation by gradually clarifying the region of interest. The red marked regions in the visualizations generated by Faster ScoreCAM and LayerCAM offer clear insights into why the model made its prediction.
In Figure \ref{fig:r4}, we observe two misclassified samples predicted by the Xception model. The XAI techniques provide valuable insights into why the model misclassified these images. Each technique highlights different regions, shedding light on the specific features that led to the misclassification. In the case of the top image, the red marked regions highlighted by the techniques are primarily located in the image border, which may not be reliable features for accurate classification. On the other hand, in the bottom image, there are no visible regions of interest, yet the visualization techniques still emphasize different regions, indicating the model's confusion in its decision-making process for this misclassification.

\section{Conclusion}\label{conclu}
The aim of this study was to classify the gastrointestinal diseases using pre-trained convolutional neural network (CNN) models. In order to enhance the accuracy of the model, we incorporated supervised contrastive learning. However, we noticed that the dataset had an imbalance issue, with certain classes having a significantly lower number of samples compared to others. To tackle this problem, we implemented cost-sensitive learning, which assigned higher weights to misclassifications in the minority class. Our experimental findings revealed a significant improvement in the classification accuracy of the CNN models through the combination of supervised contrastive learning and cost-sensitive learning. Additionally, to gain a deeper understanding of the CNN model's decision-making process, we employed explainable artificial intelligence (XAI) techniques. The use of XAI provided valuable insights into the interpretability of the CNN models by shedding light on their decision-making mechanisms. Some potential future research directions include exploring the integration of additional modalities, investigating transferability to other disease domains, developing incremental learning techniques, exploring ensemble methods for combining predictions, and tracking changes in gastrointestinal diseases for personalized treatment planning etc.
%
\bibliographystyle{splncs03_unsrt}
\bibliography{reference}

\begin{thebibliography}{10}
\providecommand{\url}[1]{\texttt{#1}}
\providecommand{\urlprefix}{URL }

\bibitem{r3}
Naz, J., Sharif, M., Yasmin, M., Raza, M., Khan, M.A.: Detection and
  classification of gastrointestinal diseases using machine learning. Current
  Medical Imaging  17(4),  479--490 (2021)

\bibitem{r4}
Borgli, H., Thambawita, V., Smedsrud, P.H., Hicks, S., Jha, D., Eskeland, S.L.,
  Randel, K.R., Pogorelov, K., Lux, M., Nguyen, D.T.D., et~al.: Hyperkvasir, a
  comprehensive multi-class image and video dataset for gastrointestinal
  endoscopy. Scientific data  7(1),  283 (2020)

\bibitem{r1}
Kaminski, M.F., Regula, J., Kraszewska, E., Polkowski, M., Wojciechowska, U.,
  Didkowska, J., Zwierko, M., Rupinski, M., Nowacki, M.P., Butruk, E.: Quality
  indicators for colonoscopy and the risk of interval cancer. New England
  journal of medicine  362(19),  1795--1803 (2010)

\bibitem{r8}
Shahzad, U., Ahmad, I., Almanjahie, I., Al-Noor, N.H.: L-moments based
  calibrated variance estimators using double stratified sampling. Computers,
  Materials \& Continua  68(3),  3411--30 (2021)

\bibitem{r9}
Abas, N., Dilshad, S., Khalid, A., Saleem, M.S., Khan, N.: Power quality
  improvement using dynamic voltage restorer. IEEE Access  8,  164325--164339
  (2020)

\bibitem{r10}
Majid, A., Khan, M.A., Yasmin, M., Rehman, A., Yousafzai, A., Tariq, U.:
  Classification of stomach infections: A paradigm of convolutional neural
  network along with classical features fusion and selection. Microscopy
  research and technique  83(5),  562--576 (2020)

\bibitem{khosla2020supervised}
Khosla, P., Teterwak, P., Wang, C., Sarna, A., Tian, Y., Isola, P., Maschinot,
  A., Liu, C., Krishnan, D.: Supervised contrastive learning. Advances in
  neural information processing systems  33,  18661--18673 (2020)

\bibitem{r13}
Jha, D., Ali, S., Tomar, N.K., Johansen, H.D., Johansen, D., Rittscher, J.,
  Riegler, M.A., Halvorsen, P.: Real-time polyp detection, localization and
  segmentation in colonoscopy using deep learning. Ieee Access  9,
  40496--40510 (2021)

\bibitem{r14}
Khan, M.A., Nasir, I.M., Sharif, M., Alhaisoni, M., Kadry, S., Bukhari, S.A.C.,
  Nam, Y.: A blockchain based framework for stomach abnormalities recognition.
  Computers, Materials \& Continua  67(1),  141--158 (2021)

\bibitem{r15}
Sun, M., Liang, K., Zhang, W., Chang, Q., Zhou, X.: Non-local attention and
  densely-connected convolutional neural networks for malignancy suspiciousness
  classification of gastric ulcer. IEEE Access  8,  15812--15822 (2020)

\bibitem{r16}
Shin, Y., Qadir, H.A., Aabakken, L., Bergsland, J., Balasingham, I.: Automatic
  colon polyp detection using region based deep cnn and post learning
  approaches. IEEE Access  6,  40950--40962 (2018)

\bibitem{r17}
Ribeiro, J., N{\'o}brega, S., Cunha, A.: Polyps detection in colonoscopies.
  Procedia Computer Science  196,  477--484 (2022)

\bibitem{r23}
He, K., Zhang, X., Ren, S., Sun, J.: Deep residual learning for image
  recognition. In: Proceedings of the IEEE conference on computer vision and
  pattern recognition. pp. 770--778 (2016)

\bibitem{r24}
Tan, M., Le, Q.: Efficientnet: Rethinking model scaling for convolutional
  neural networks. In: International conference on machine learning. pp.
  6105--6114. PMLR (2019)

\bibitem{r25}
Huang, G., Liu, Z., Van Der~Maaten, L., Weinberger, K.Q.: Densely connected
  convolutional networks. In: Proceedings of the IEEE conference on computer
  vision and pattern recognition. pp. 4700--4708 (2017)

\bibitem{r26}
Szegedy, C., Liu, W., Jia, Y., Sermanet, P., Reed, S., Anguelov, D., Erhan, D.,
  Vanhoucke, V., Rabinovich, A.: Going deeper with convolutions. In:
  Proceedings of the IEEE conference on computer vision and pattern
  recognition. pp. 1--9 (2015)

\bibitem{r27}
Chollet, F.: Xception: Deep learning with depthwise separable convolutions. In:
  Proceedings of the IEEE conference on computer vision and pattern
  recognition. pp. 1251--1258 (2017)

\bibitem{gradcam}
Selvaraju, R.R., Cogswell, M., Das, A., Vedantam, R., Parikh, D., Batra, D.:
  Grad-cam: Visual explanations from deep networks via gradient-based
  localization. In: Proceedings of the IEEE international conference on
  computer vision. pp. 618--626 (2017)

\bibitem{gradcamplusplus}
Chattopadhay, A., Sarkar, A., Howlader, P., Balasubramanian, V.N.: Grad-cam++:
  Generalized gradient-based visual explanations for deep convolutional
  networks. In: 2018 IEEE winter conference on applications of computer vision
  (WACV). pp. 839--847. IEEE (2018)

\bibitem{scorecam}
Wang, H., Wang, Z., Du, M., Yang, F., Zhang, Z., Ding, S., Mardziel, P., Hu,
  X.: Score-cam: Score-weighted visual explanations for convolutional neural
  networks. In: Proceedings of the IEEE/CVF conference on computer vision and
  pattern recognition workshops. pp. 24--25 (2020)

\bibitem{layercam}
Jiang, P.T., Zhang, C.B., Hou, Q., Cheng, M.M., Wei, Y.: Layercam: Exploring
  hierarchical class activation maps for localization. IEEE Transactions on
  Image Processing  30,  5875--5888 (2021)

\bibitem{lime}
Ribeiro, M.T., Singh, S., Guestrin, C.: "why should i trust you?" explaining
  the predictions of any classifier. In: Proceedings of the 22nd ACM SIGKDD
  international conference on knowledge discovery and data mining. pp.
  1135--1144 (2016)

\bibitem{r11}
PH, B.H.T.V.S., Hicks, S., et~al.: Hyperkvasir, a comprehensive multi-class
  image and video dataset for gastrointestinal endoscopy sci. Data  7(1), ~1
  (2020)

\bibitem{9850571}
Gupta, D., Anand, G., Kirar, P., Meel, P.: Classification of endoscopic images
  and identification of gastrointestinal diseases. In: 2022 International
  Conference on Machine Learning, Big Data, Cloud and Parallel Computing
  (COM-IT-CON). vol.~1, pp. 231--235 (2022)

\bibitem{9474707}
Gjestang, H.L., Hicks, S.A., Thambawita, V., Halvorsen, P., Riegler, M.A.: A
  self-learning teacher-student framework for gastrointestinal image
  classification. In: 2021 IEEE 34th International Symposium on Computer-Based
  Medical Systems (CBMS). pp. 539--544 (2021)

\end{thebibliography}

\end{document}